\documentclass{article}
\usepackage{ijcai24}

\usepackage{times}
\usepackage{soul}
\usepackage{url}
\usepackage[hidelinks]{hyperref}
\usepackage[utf8]{inputenc}
\usepackage[small]{caption}
\usepackage{graphicx}
\usepackage{amsmath}
\usepackage{amsthm}
\usepackage{booktabs}
\usepackage[switch]{lineno}

\usepackage{amsmath,amssymb,amsfonts}
\usepackage{subfigure}
\usepackage{textcomp}
\usepackage{bm}
\usepackage{xcolor}
\usepackage{float}

\usepackage{multirow}
\usepackage{makecell}
\usepackage{array}

\usepackage[ruled,vlined]{algorithm2e}

\usepackage{colortbl} 

\usepackage{marvosym} 

\title{PTDE: Personalized Training with Distilled Execution for Multi-Agent Reinforcement Learning}

\author{
Yiqun Chen$^1$\and
Hangyu Mao$^2$\and
Jiaxin Mao$^1$\thanks{Corresponding author}\Letter \and
Shiguang Wu$^3$\and 
Tianle Zhang$^4$\and \\
Bin Zhang$^5$\and
Wei Yang$^5$\and
Hongxing Chang$^5$\\
\affiliations
$^1$Renmin University of China, 
$^2$SenseTime, 
$^3$Noah's Ark Lab, Huawei, \\
$^4$JD Explore Academy, 
$^5$Institute of Automation,Chinese Academy of Sciences\\
\emails
chenyiqun990321@ruc.edu.cn,
hy.mao@pku.edu.cn,
\{maojiaxin, weiyangvia\}@gmail.com,
wushiguang@huawei.com,
tianle-zhang@outlook.com,
\{zhangbin2020, hongxing.chang\}@ia.ac.cn
}

\begin{document}

\maketitle

\begin{abstract}
Centralized Training with Decentralized Execution (CTDE) has emerged as a widely adopted paradigm in multi-agent reinforcement learning, emphasizing the utilization of global information for learning an enhanced joint $Q$-function or centralized critic. In contrast, our investigation delves into harnessing global information to directly enhance individual $Q$-functions or individual actors. Notably, we discover that applying identical global information universally across all agents proves insufficient for optimal performance. Consequently, we advocate for the customization of global information tailored to each agent, creating agent-personalized global information to bolster overall performance. Furthermore, we introduce a novel paradigm named Personalized Training with Distilled Execution (PTDE), wherein agent-personalized global information is distilled into the agent's local information. This distilled information is then utilized during decentralized execution, resulting in minimal performance degradation. PTDE can be seamlessly integrated with state-of-the-art algorithms, leading to notable performance enhancements across diverse benchmarks, including the SMAC benchmark, Google Research Football (GRF) benchmark, and Learning to Rank (LTR) task.
\end{abstract}

\section{Introduction}

Many real-world tasks can be modeled as decision problems for multi-agent systems, such as multi-robot navigation ~\cite{navigation}, multi-robot collision avoidance \cite{collision-avoidance}, multi-UAV path planning \cite{uav}, information retrieval \cite{chen2024ma4div} and games \cite{mao2021seihai,guss2021towards}. In most of these scenarios, the prevalent constraints include partial observability, and agents are constrained to decentralized decision-making processes.

To address these challenges, Multi-Agent Reinforcement Learning (MARL) has emerged as a focal point of research. Within MARL, Centralized Training with Decentralized Execution (CTDE) stands out as a prominent paradigm. CTDE leverages global information during training and shifts to utilizing only local information during execution, facilitating decentralized decision-making. This paradigm encompasses two primary algorithmic categories: value-decomposition based and actor-critic based approaches. Concerning the utilization of global information, the former category \cite{VDN,QMIX,QPLEX,UNMAS} employs global information for enhancing the joint $Q$-function. In contrast, the latter category \cite{MADDPG,COMA,MAAC,MAPPO,STEP} incorporates global information as input to a centralized critic. Notably, these approaches refrain from utilizing global information directly during execution, a factor that could potentially constrain collaboration performance among agents, especially in complex scenarios, as demonstrated in our experiments.

In contrast to the conventional CTDE approach, a distinct line of research explores the direct utilization of global information during execution. COPA \cite{COPA} introduces a Coach-Player framework, devising an adaptive communication method wherein the coach determines when to dispatch a global instruction vector to the players. This vector, combined with local information, is used to compute individual $Q$-functions. Despite COPA's use of a multi-head attention mechanism for comprehensive global information processing during execution, this information remains identical for all agents. In a different vein, the CSRL framework \cite{CSRL} introduces a Commander-Soldiers MARL framework, incorporating the concept of specific information for each agent. Both COPA and CSRL notably enhance multi-agent collaboration performance by directly applying global information during execution. 


Nevertheless, practical challenges emerge in numerous applications due to local observability constraints, posing difficulties in utilizing global information directly during execution. To reconcile the need for global information while ensuring decentralized execution, we propose a novel paradigm named Personalized Training with Distilled Execution (PTDE), which comprises two training stages. In the first stage, we introduce the concept of \textit{agent-personalized global information} by employing a Global Information Personalization (GIP) module. This module transforms raw global information into personalized global information tailored to each agent. This personalized global information is then utilized to compute individual $Q$-functions or individual policies, enhancing the performance of each agent. In the second stage, we implement knowledge distillation for the agent-personalized global information. Within this distillation framework, a proficiently trained GIP module acts as the teacher network, while a dedicated student network is employed for the distillation process. Crucially, the input for the student network is exclusively composed of the agent's local information, presenting a departure from the teacher network, which integrates both global and agent's local information \footnote{Our approach diverges from the conventional knowledge distillation employed in model compression \cite{MC}, where both the teacher and student networks operate on the same input.}. 

During execution, the teacher network is replaced by the student network, enabling decentralized execution while retaining the benefits of personalized global information. This innovative approach ensures a seamless transition from personalized training to distilled execution within the proposed PTDE paradigm. Summary of our contributions:



\begin{itemize}
\item In contrast to the prevalent trend in CTDE-based methods, which emphasizes leveraging global information during centralized training, our approach shifts the focus to exploring the utilization of global information during decentralized execution.
\item We identify that consistently positive performance among agents is challenging when applying the same global information for decision-making. However, our innovation lies in transforming global information into agent-personalized global information, resulting in agents consistently making improved decisions.
\item We introduce a novel paradigm named PTDE, which not only benefit from agent-personalized global information but also executes in a decentralized manner through knowledge distillation. Importantly, our experiments demonstrate minimal performance degradation after distilling agent-personalized global information into agent’s local information. 
\item Experimental results underscore the universality and efficacy of the PTDE paradigm across diverse multi-agent environments and algorithms.

\end{itemize}

\section{Background}
\subsection{Dec-POMDP}
In this work, we model a fully cooperative multi-agent task as the Dec-POMDP \cite{DECPOMDP}, which is formally defined as a tuple $G = \left \langle \textbf{S}, \textbf{U}, P, r, \textbf{Z}, \textbf{O}, n, \gamma \right \rangle $. $s \in \textbf{S}$ is the global state of the environment. Each agent $i \in \mathcal{A} \equiv \{1, ..., n\}$ chooses an action $u^i \in U$ which forms the joint action $\textbf{u} \in \textbf{U} \equiv U^n$. The state transition function is modeled as $P(s^{\prime}|s, \textbf{u}): \textbf{S} \times \textbf{U} \times \textbf{S} \to [0, 1]$. The reward function which is modeled as $r(s, \textbf{u}): \textbf{S} \times \textbf{U}$ is shared by all agents and the discount factor is $\gamma \in [0, 1)$. It follows partially observable settings, where agents do not have access to the global state. Instead, it samples observations $z \in \textbf{Z}$ according to observation function $O(s, i): \textbf{S} \times \textbf{U} \to \textbf{Z}$. Each agent has an action-observation history trajectory $\tau^i \in T \equiv (\textbf{Z} \times \textbf{U})^*$, on which it conditions a stochastic policy $\pi^i(u^i|\tau^i): T \times \textbf{U} \to [0, 1]$. In our algorithm, the joint policy $\pi$ is based on action-value function $Q^{\pi}(s_t, \textbf{u}_t) = \mathbb{E}_{s_{t+1}:\infty,\textbf{u}_{t+1}:\infty}[\sum_{k=0}^{\infty}\gamma^k r_{t+k} | s_t, \textbf{u}_t]$. The final goal is to get the optimal action-value function $Q^*$.

\subsection{Typical MARL Algorithms}

Value decomposition \cite{VDN,QMIX,QPLEX} and Actor-Critic \cite{MAPPO,zhang2022efficient,zhang2023stackelberg,hu2024measuring} are two typical branches of multi-agent reinforcement learning. Among these, VDN \cite{VDN} is the representative algorithm to formulate value-decomposition paradigm. QMIX \cite{QMIX} learns a monotonic factorisation ensuring that a global argmax operation on the joint action-value function $Q_{tot}$ yield the same results as a series of individual argmax operations on each individual action-value function $Q_i$. QPLEX \cite{QPLEX} takes a duplex dueling network architecture to factorize the joint value function. MAPPO \cite{MAPPO} is an actor-critic based algorithm. To specialize for multi-agent settings, MAPPO uses the structure of PPO algorithm but the critic can take extra global information to follow the CTDE framework.

\subsection{Knowledge Distillation}

Knowledge distillation \cite{KD} is proposed to compress big models. It distills the knowledge generated from a larger network into a smaller network. Policy distillation \cite{PD} presents a novel knowledge distillation method which can be used in reinforcement learning to extract the policy of agent and train a new network with an expert level performance and better efficiency. CTDS \cite{CTDS} proposes a novel Centralized Teacher with Decentralized Student framework which consists of a teacher model and a student model to alleviate the inefficiency caused by the limitation of local observability. 


\section{Method}
In this section, we initially present an approach that provides the same global information to all agents during execution for decision-making. Despite its simplicity, this naive use of global information does not consistently enhance multi-agent collaboration performance. Subsequently, we propose the Global Information Personalization (GIP) module to tailor global information for each agent, resulting in the agent-personalized global information. Based on this, we derive a centralized execution method that makes better use of global information for improved performance. Recognizing the challenge of directly obtaining global information during execution, we finally introduce the knowledge distillation approach to achieve decentralized execution without too much performance degradation.  

\subsection{Naive Use of Global Information}

\begin{figure}[htbp]
\centerline{\includegraphics[width=0.4\textwidth]{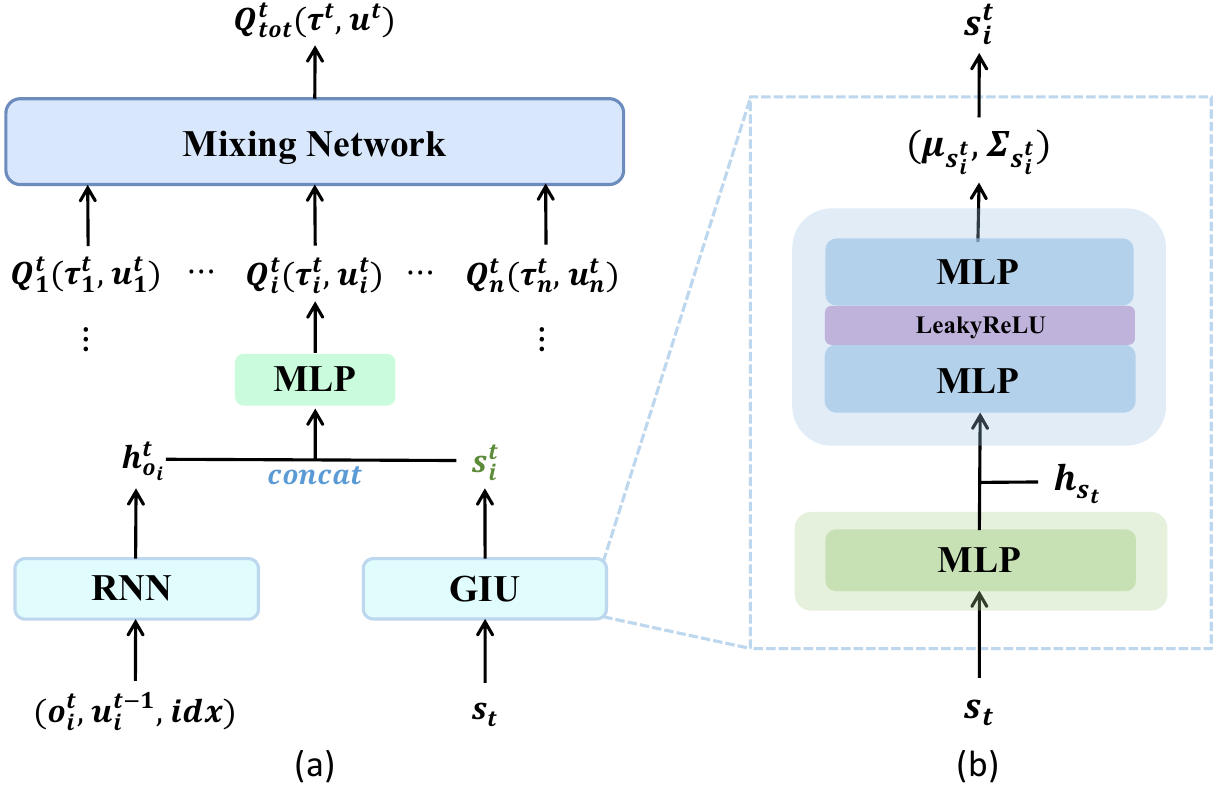}}
\caption{The framework of QMIX\_GIU. (b) is the detail of the Global Information Unification (GIU) module.}
\label{QMIX_GIU}
\end{figure}

We present a method that directly employs unified global information for all agents during execution. Using QMIX as an example, the entire framework is depicted in Figure \ref{QMIX_GIU} (a).

(1) The RNN module encodes the trajectory of the \emph{agent's local information} $\mathbf{O_i^t} = (o_i^t, u_i^{t-1}, idx)$ to $h_{o_i}^t$. 

(2) The Global Information Unification (GIU) module is designed to generate the \emph{unified global information} $s_i^t$ to be used in execution, where the green multilayer perceptron (MLP) encodes the \emph{raw global information} $s_t$ into $h_{s_t}$ and the blue module transforms $h_{s_t}$ into a multivariate gaussian distribution $\mathcal N \sim  (\bm{\mu_{s_i^t}}, \bm{\Sigma_{s_i^t}})$. 

(3) Similar to QMIX, the individual action-value $Q_i^t(\tau_i^t, u_i^t)$ is computed by an MLP operating on the concatenation of $h_{o_i}^t$ and $s_i^t$, while the joint action-value $Q_{tot}^t(\tau^t, u^t)$ is computed by nonlinearly combining all individual action-values through the mixing network.

Since the parameters of GIU module are unified and invariant to each agent during execution, the algorithm is called QMIX\_GIU (\textbf{G}lobal \textbf{I}nformation \textbf{U}nification). It is one of the baselines and ablations in our experiments.

\subsection{Global Information Personalization}

In many multi-agent cooperative tasks, an agent's decision-making is significantly improved by concentrating on a subset of the global information, given that the raw global information tends to be redundant \cite{mao2020learning}. Extracting and utilizing this relevant portion of global information is crucial for optimal decision-making. Inspired by this observation, we introduce the Global Information Personalization (GIP) module, crafted to autonomously tailor the global state (i.e., extracting the beneficial part) for each individual agent.

\begin{figure}[htbp]
\centerline{\includegraphics[width=0.4\textwidth]{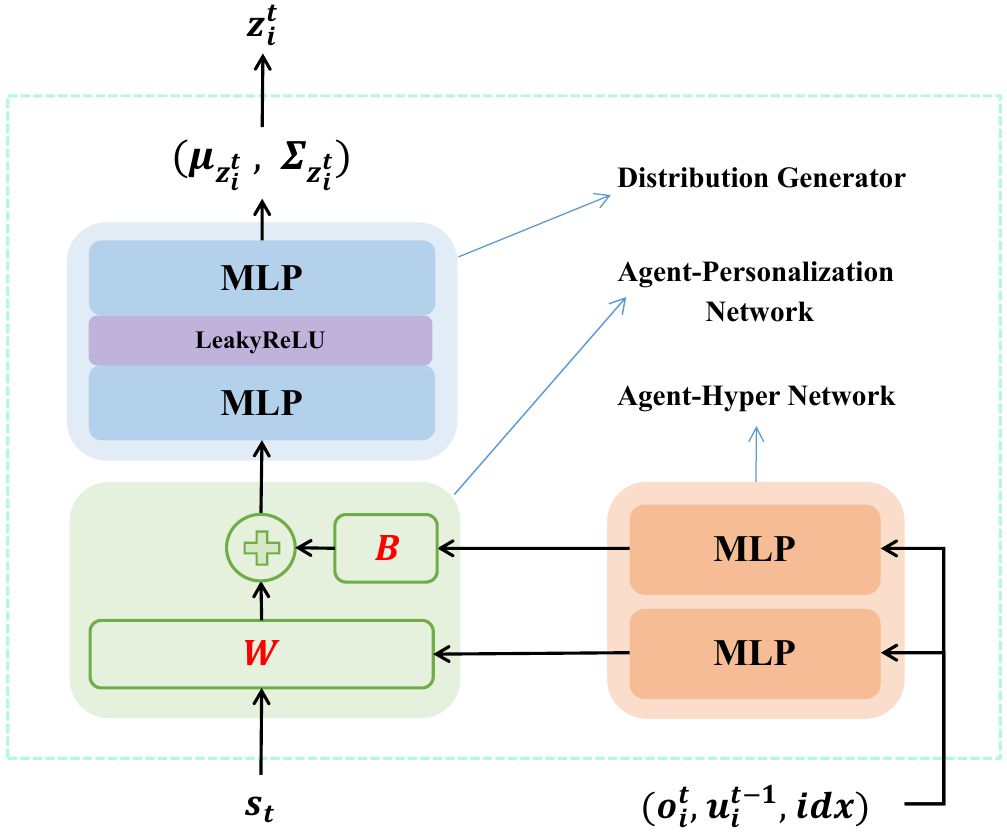}}
\caption{The structure of the Global Information Personalization (GIP) module.}
\label{GISM}
\end{figure}

As shown in Figure \ref{GISM}, the GIP module comprises three components: Agent-Hyper Network, Agent-Personalization Network, and Distribution Generator. The Agent-Hyper Network takes the agent's local information as input and produces a set of weights $W$ and biases $B$. The structures of the Agent-Personalization Network and Distribution Generator are identical to those in Figure \ref{QMIX_GIU} (b). The output of the GIP module, denoted as $z_i^t$, is defined by Equation (\ref{zitN}).


\begin{equation}
z_i^t \sim \mathcal N (\bm{\mu_{z_i^t}}, \bm{\Sigma_{z_i^t}})
\label{zitN}
\end{equation}

\vspace{-0.15cm}

\begin{equation}
\bm{\mu_{z_i^t}} = f_{\mu}(\mathbf{O_i^t}, \mathbf{s}; \theta_{\mu})
\label{zitN2}
\end{equation}

\vspace{-0.15cm}

\begin{equation}
\bm{\Sigma_{z_i^t}} = f_{\Sigma}(\mathbf{O_i^t}, \mathbf{s}; \theta_{\Sigma})
\label{zitN3}
\end{equation}

Compared to the GIU module (i.e., Figure \ref{QMIX_GIU} (b)), a distinguishing feature of GIP module is that the parameters of Agent-Personalization Network are dynamically generated by Agent-Hyper Network. Since the local information is different for each agent, the parameters of Agent-Personalization Network are guaranteed to be personalized to each agent, which is the key to global information personalization. Therefore, we call $z_i^t$ the \emph{agent-personalized global information} in this paper. 

\begin{figure}[htbp]
\centering
\subfigure[GIP\_Q]{
\begin{minipage}[t]{0.15\textwidth}
\includegraphics[width=1\textwidth]{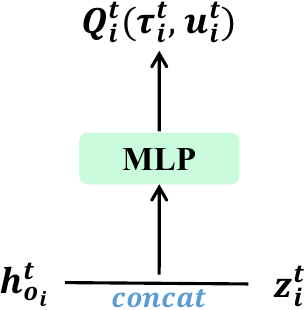}
\end{minipage}
}
\hspace{1cm}
\subfigure[GIP\_AC]{
\begin{minipage}[t]{0.15\textwidth}
\includegraphics[width=1\textwidth]{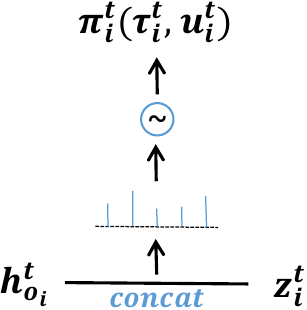}
\end{minipage}
}
\caption{How GIP module is used in value-decomposition based methods (i.e., GIP\_Q) and actor-critic based methods (i.e., GIP\_AC).}
\label{GISM_Q_AC}
\end{figure}

The GIP module is versatile across existing CTDE algorithms. In Figure \ref{GISM_Q_AC} (a), the integration of the GIP module in value-decomposition methods is illustrated. Here, the individual $Q$-function is computed by concatenating $h_{o_i}^t$ and $z_i^t$. Figure \ref{GISM_Q_AC} (b) showcases the application of the GIP module in actor-critic methods, where the concatenation $[h_{o_i}^t, z_i^t]$ is utilized by the individual actor for action sampling.

\subsection{Knowledge Distillation}
The methods shown in Figure \ref{GISM_Q_AC} involve the utilization of the global state $s_t$ when computing individual $Q$-functions or individual policies (as $z_i^t$ depends on $s_t$). However, obtaining global information directly is challenging due to partial observability in real-world multi-agent systems. To leverage global information during execution while adhering to the need for decentralized execution, we introduce a knowledge distillation method. This approach distills agent-personalized global information using only the agent's local information, i.e., transforming the dependence of $z_i^t$ on $s_t$ into the practical reliance on $(o_i^t, u_i^{t-1}, idx)$.

\begin{figure}[htbp]
\centerline{\includegraphics[width=0.45\textwidth]{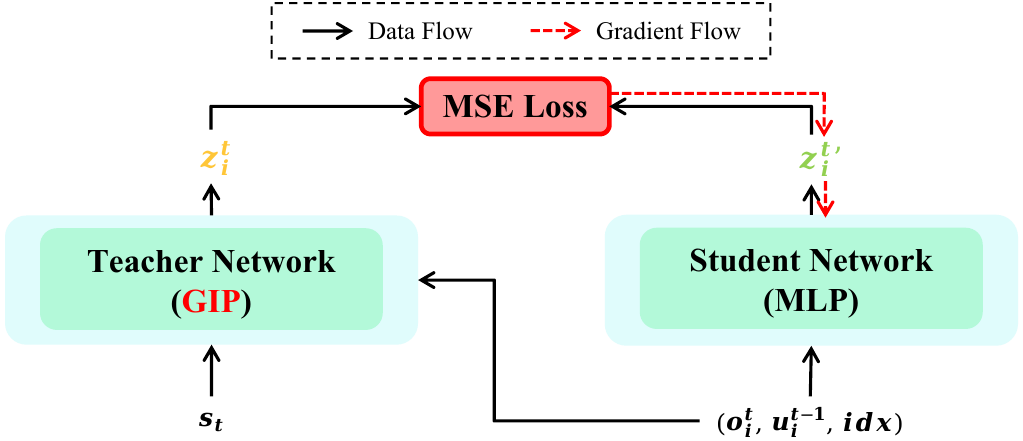}}
\caption{The knowledge distillation framework.}
\label{GKD}
\end{figure}

The knowledge distillation framework is illustrated in Figure \ref{GKD}. In this setup, the GIP module serves as the teacher network, while the student network is represented by an MLP. As indicated in Equation (\ref{zit}), the calculation of $z_i^t$ involves $s_t$ and $\mathbf{O_i^t} = (o_i^t, u_i^{t-1}, idx)$ through the teacher network. On the other hand, the student network's input consists solely of the agent's local information $O_i^t$, and its output is denoted as ${z_i^t}'$ in Equation (\ref{zit'}). In the context of knowledge distillation, $z_i^t$ is referred to as teacher knowledge, and ${z_i^t}'$ is student knowledge. Equation (\ref{MSE}) illustrates the use of Mean Squared Error (MSE) Loss to minimize the disparity between student and teacher knowledge, ensuring effective training of the student network. 



\begin{equation}
  \begin{aligned}
  z_i^t = f_{tea}(\mathbf{O_i^t}, s_t) \sim \mathcal N \left(\bm{\mu_{z_i^t}}(\mathbf{O_i^t}, s_t), \bm{\Sigma_{z_i^t}}(\mathbf{O_i^t}, s_t) \right)
  \end{aligned}
  \label{zit}
\end{equation}



\begin{equation}
  \begin{aligned}
  {z_i^t}' &= f_{stu}(\mathbf{O_i^t}) = \mathbf{MLP(O_i^t)}
  \end{aligned}
  \label{zit'}
\end{equation}


\begin{equation}
  \begin{aligned}
  \mathcal{L}_{mse} &= ||{z_i^t}' - z_i^t||_{2}^{2} = ||f_{tea}(\mathbf{O_i^t}, s_t) - f_{stu}(\mathbf{O_i^t})||_{2}^{2}
  \end{aligned}
  \label{MSE}
\end{equation}

Knowledge distillation is a common technique employed for model compression, typically involving identical input data for both teacher and student networks. In our knowledge distillation, a notable distinction arises as the student network's input lacks global information $s_t$ compared to the teacher network. This divergence from traditional model compression is pivotal, serving as the key factor in transitioning from centralized execution to decentralized execution.

Upon completion of the knowledge distillation training, the student network can seamlessly replace the teacher network (i.e., the GIP module) during the execution process. In other words, ${z_i^t}'$ is utilized in lieu of $z_i^t$ in Figure \ref{GISM_Q_AC}, enabling decentralized execution.

\subsection{The Overall PTDE Paradigm}


\begin{figure*}[htbp]
\centering
\subfigure[Two-stage training]{
\begin{minipage}[t]{0.63\textwidth}
\includegraphics[width=1\textwidth]{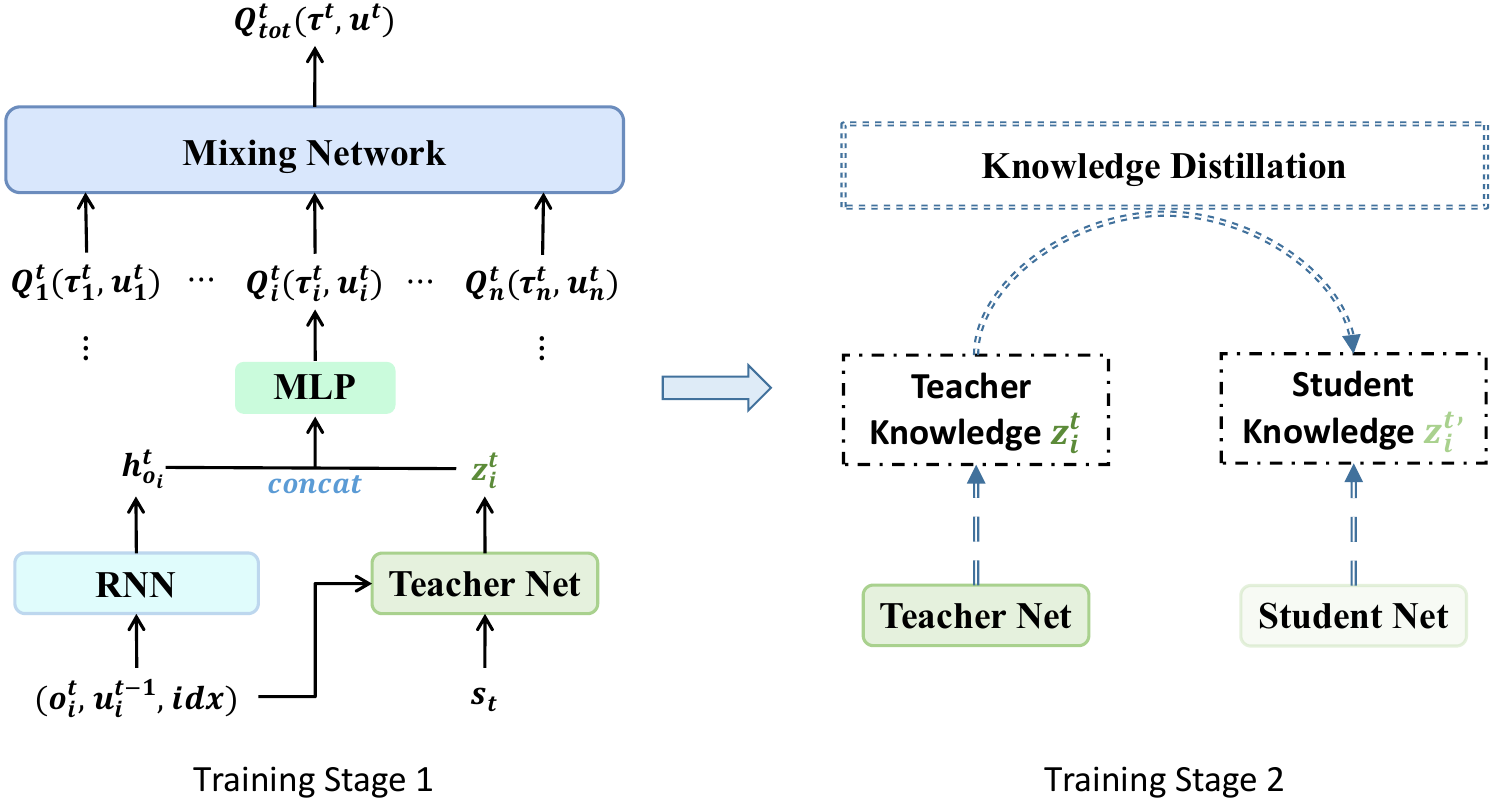}
\end{minipage}
}
\subfigure[Decentralized Execution]{
\begin{minipage}[t]{0.24\textwidth}
\raisebox{0.5cm}{\includegraphics[width=1\textwidth]{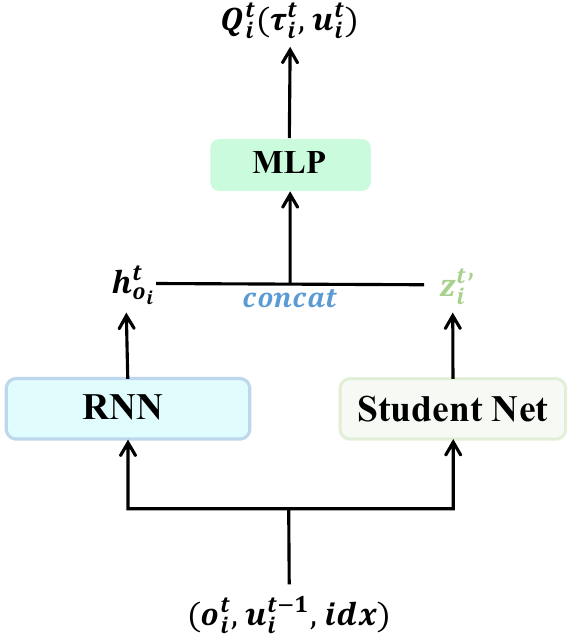}}
\end{minipage}
}
\caption{The framework of PTDE: Two-Stage Training and Decentralized Execution.}
\label{training and execution}
\end{figure*}

Figure \ref{training and execution} shows the overall framework of PTDE based on the QMIX algorithm, encompassing two-stage training and decentralized execution.

\textbf{The First Training Stage.} We provide personalized global information for each agent to compute a better individual $Q$-function or individual policy. Specifically, the RNN module records the trajectory of agent's local information and outputs an encoding vector $h_{o_i}^t$. The teacher network gives agent-personalized global information $z_i^t$. Then, the individual action-value $Q_i^t(\tau_i^t, u_i^t)$ can be computed by concatenation $[h_{o_i}^t, z_i^t]$. Finally, the joint action-value $Q_{tot}^t(\bm{\tau^t}, \bm{u^t})$ can be obtained by the mixing network. The whole network is trained end-to-end by minimizing the loss shown in Equation (\ref{TD loss}) and (\ref{target y}):


\begin{equation}
\mathcal{L}(\theta, \varphi) = \sum_{i=1}^b \left(y_i^{tot}-Q_{tot}(\bm{\tau}, \textbf{u}, s; \theta, \varphi) \right)^2
\label{TD loss}
\end{equation}


\begin{equation}
y^{tot} = r + \gamma \max_{u'} Q_{tot}(\bm{\tau}', \textbf{u}', s'; \theta^{-}, \varphi^{-})
\label{target y}
\end{equation}

where $b$ is the batch size of sampled experiences from replay buffer; $\varphi$ is the parameters of the teacher network and $\theta$ is parameters of the other networks in the first training stage; $\theta^{-}$ and $\varphi^{-}$ are parameters of target networks. We also show the pseudo-code in \textbf{Algorithm 1}.


    
    
    
      


\begin{algorithm}
  \DontPrintSemicolon
  \SetKwInOut{KwOutput}{Output}

  \textbf{Initialize}: The parameters $\theta$ and $\varphi$ of network, $\theta^{-}$ and $\varphi^{-}$ of target network, replay buffer $\mathcal{D}$.\\
  \textbf{Initialize}: Observation $\bm{o}$ $=(o_1,\cdots,o_N)$ and state $\bm{s}$.\\
  \While{not over}{
    Collect a tuple ($\bm{o}, \bm{s}, \bm{a}, \bm{r}, \bm{o^{\prime}}, \bm{s^{\prime}}$) by generating 8 parallel episodes, and store it in $\mathcal{D}$;\\
    Sample a random minibatch $\bm{b}$ from $\mathcal{D}$;\\
    The RNN calculate $h_{o_i}^t$, the teacher network calculate $z_i^t$;\\
    Calculate $y^{tot}$ and loss $\mathcal{L}$ for all sampled data from $\bm{b}$ based on Equation (\ref{target y}) and (\ref{TD loss});\\
    Update the parameters of networks $\theta$, $\varphi$;\\
    Update the parameters of target networks $\theta^{-}$, $\varphi^{-}$ every $N$ episodes;
  }
  \textbf{Output}: Get a well-trained teacher network and an algorithm that works well to execute centrally.
  \caption{The first training stage of PTDE}
\end{algorithm}


\textbf{The Second Training Stage.} After the first training stage, the teacher network can provide agent-personalized global knowledge $z_i^t$ for agents' decision-making. In the second stage, we use offline knowledge distillation where the student network distills teacher knowledge $z_i^t$ to obtain student knowledge ${z_i^t}'$. Subsequently, the student network can replace the teacher network during the execution process. The pseudo-code can be seen in \textbf{Algorithm 2}. 


      

\begin{algorithm}
\DontPrintSemicolon
    \textbf{Initialize}: The student network parameters $\psi$. \\
    \textbf{Load Model}: Load the models and parameters $\theta$ and $\varphi$ in Algorithm 1. \\
    \textbf{Generate Data}: Generate 100 episodes data, including $s_t$ and $(o_i^t, u_i^{t-1}, idx)$, and save offline. \\
    \textbf{Offline Train}: Use MSE Loss in Equation (\ref{MSE}) to train the student network offline for multiple epochs. \\
    \textbf{Output}: Get a well-trained student network.
    \caption{The second training stage of PTDE}
\end{algorithm}

\textbf{Decentralized Execution.} As shown in Figure \ref{training and execution} (b), agents utilize solely local information $(o_i^t, u_i^{t-1}, idx)$ to compute  individual action-values $Q_i^t(\tau_i^t, u_i^t)$ and sample actions using  $u_i^t = \arg\max_{u} Q_i^t(\tau_i^t, u_i^t)$. We name this method as QMIX\_KD.



\section{Experiments}

Our experiments mainly focus on six research questions: 

(\textbf{RQ.1}) Does the utilization of unified global information during execution lead to an enhancement in the performance of multi-agent collaboration?

(\textbf{RQ.2}) Is agent-personalized global information more effective in improving performance compared to unified global information?

(\textbf{RQ.3}) After knowledge distillation, can the algorithm maintain a substantial portion of its performance when transitioning from centralized to decentralized execution?

(\textbf{RQ.4}) Does the PTDE-based algorithm exhibit universality across diverse environments?

(\textbf{RQ.5}) Does the PTDE paradigm demonstrate universality across various algorithms?

(\textbf{RQ.6}) What is the rationale behind the PTDE paradigm's approach of partitioning the training process into two stages?

We investigate the research questions using popular MARL testbeds, namely StarCraft II \cite{SMAC} and Google Research Football \cite{GRF}. Additionally, we validate RQ.4 through experiments in the Learning to Rank (LTR) scenario \cite{LTR}. To our best knowledge, this is the first time that the MARL algorithm has been applied to LTR tasks.


For baselines, we categorize them into two classes: centralized execution algorithms and decentralized execution algorithms, outlined in Table \ref{baselines}. To showcase the impact of agent-personalized global information, we contrast our approach with two centralized execution algorithms, CSRL and COPA, and perform an ablation experiment using QMIX\_GIU. Furthermore, to assess algorithm performance after knowledge distillation, we utilize QMIX and QPLEX as decentralized execution baselines. All experiments are conducted using the PyMARL2 framework \cite{PYMARL2} with 8 parallel runners and 3 random seeds. Details regarding hyperparameters are available in Table \ref{hyper parameters} in the Appendix.

\begin{table}[] \scriptsize
\centering
\caption{Algorithms and baselines in experiments.}
\begin{tabular}{c|c|c}
\hline
                                         & Algorithm  & Description                                                          \\ \hline
\multirow{4}{*}{\makecell[c]{Centralized \\ Execution}}   & CSRL       &  \cite{CSRL}                                                                    \\ \cline{2-3} 
                                         & COPA       & \cite{COPA}                                                                     \\ \cline{2-3} 
                                         & QMIX\_GIU  & \makecell[c]{Unified global information\\ is used during execution. \\ (Proposed in Section 3.1)}              \\ \cline{2-3} 
                                         & QMIX\_GIP (stage1)  & \makecell[c]{Agent-Personalized global\\ information is used during\\ execution.}              \\ \hline
\multirow{3}{*}{\makecell[c]{Decentralized \\ Execution}} & QMIX\_KD (stage2) & \makecell[c]{Student knowledge (rather than \\ teacher knowledge) is used\\ during execution.} \\ \cline{2-3} 
                                         & QMIX       &  \cite{QMIX}                                                                    \\ \cline{2-3} 
                                         & QPLEX      &  \cite{QPLEX}                                                                    \\ \hline
\end{tabular}
\centering
\label{baselines}
\end{table}

\subsection{StarCraft II}

\begin{table*}[]\scriptsize 
\centering\caption{Winning rates on StarCraft II.}
\renewcommand{\arraystretch}{1.08} 
\begin{tabular}{lc|c@{\hspace{1mm}}c@{\hspace{1mm}}c@{\hspace{1mm}}c@{\hspace{1mm}}c@{\hspace{1mm}}c}
\hline
\multicolumn{1}{c|}{}                                         & Algorithms & \emph{3s\_vs\_5z} (2M)  & \emph{5m\_vs\_6m} (2M)  & \emph{3s5z\_vs\_3s6z} (5M) & \emph{6h\_vs\_8z} (5M)  & \emph{3s\_vs\_8z} (10M)  & \emph{3s5z\_vs\_3s7z} (10M) \\ \hline
\multicolumn{1}{c|}{\multirow{4}{*}{\makecell[c]{Centralized \\ Execution}}} & CSRL    & 0.917±0.031 & 0.733±0.063 & 0.396±0.267    & 0.502±0.373 & 0.934±0.019 & 0.109±0.092    \\ 
\multicolumn{1}{c|}{}                                         & COPA    & 0.748±0.087 & 0.688±0.117 & 0.064±0.098    & 0.709±0.129 & 0.490±0.230 & 0.000±0.000    \\ 
\multicolumn{1}{c|}{}                                         & QMIX\_GIU    & 0.868±0.095 & 0.696±0.079 & 0.026±0.040    & 0.405±0.310 & 0.332±0.264 & 0.000±0.000    \\ 
\multicolumn{1}{c|}{}                                         & \textbf{QMIX\_GIP (stage1)}    & \textbf{0.992±0.006} & \textbf{0.806±0.008} & \textbf{0.776±0.062} & \textbf{0.712±0.053} & \textbf{0.990±0.002} & \textbf{0.710±0.152} \\ \hline
\multicolumn{1}{c|}{\multirow{3}{*}{\makecell[c]{Decentralized \\ Execution}}}   & \textbf{QMIX\_KD (stage2)} & \textbf{0.887±0.027} & \textbf{0.690±0.088} & \textbf{0.674±0.069} & \textbf{0.524±0.085} & \textbf{0.576±0.055} & \textbf{0.631±0.201} \\ 
\multicolumn{1}{c|}{}                                         & QMIX         & 0.128±0.165 & 0.586±0.068 & 0.140±0.081    & 0.012±0.019 & 0.355±0.197 & 0.000±0.000    \\ 
\multicolumn{1}{c|}{}                                         & QPLEX        & 0.000±0.000 & 0.616±0.070 & 0.390±0.145    & 0.021±0.034 & 0.074±0.066 & 0.000±0.000    \\ \hline
\multicolumn{2}{c|}{Performance Retention Ratio (PRR)}   &  89.4\% &   85.6\%  &   86.9\%  &  73.6\%   &   58.2\%   &  88.9\%       \\ \hline
\end{tabular}
\label{starcraft 2 winning rate}
\end{table*}

To address \textbf{RQ.1}, \textbf{RQ.2}, and \textbf{RQ.3}, we select hard scenarios such as \emph{5m\_vs\_6m} and \emph{3s\_vs\_5z}, as well as super hard scenarios like \emph{3s5z\_vs\_3s6z} and \emph{6h\_vs\_8z}. Additionally, to further highlight the advantages of agent-personalized global information in multi-agent collaboration, we conduct experiments in two new scenarios, namely \emph{3s\_vs\_8z} (featuring 8 zealots in the enemy team) and \emph{3s5z\_vs\_3s7z} (with 3 stalkers and 7 zealots in the enemy team), where existing decentralized execution algorithms exhibit poor performance.

As shown in Table \ref{starcraft 2 winning rate}, QMIX\_GIU has better performance than QMIX in \emph{3s\_vs\_5z}, \emph{5m\_vs\_6m} and \emph{6h\_vs\_8z}, but performs worse in \emph{3s\_vs\_8z} and \emph{3s5z\_vs\_3s6z}. This addresses \textbf{RQ.1}, indicating that unified global information can have a positive impact on decision-making in certain scenarios but may not consistently improve agent decisions. In contrast, QMIX\_GIP consistently outperforms QMIX\_GIU in all testing scenarios, supporting \textbf{RQ.2} by highlighting the consistent benefits of agent-personalized global information. Furthermore, QMIX\_GIP attains the highest winning rates among all centralized execution algorithms, further affirming the advantages of agent-personalized global information (\textbf{RQ.2}). 

The Performance Retention Ratio (PRR) in Table \ref{starcraft 2 winning rate} signifies the ratio of the winning rates of QMIX\_KD to those of QMIX\_GIP, reflecting the performance before and after knowledge distillation. The PRRs range between 85\% and 90\% in four out of six simulation maps, indicating that the PTDE paradigm can maintain performance reasonably well after knowledge distillation (\textbf{RQ.3}). Notably, even in challenging scenarios like \emph{3s5z\_vs\_3s7z}, where all baselines have low winning rates, QMIX\_KD achieves a winning rate of 63.1\%, showcasing the substantial advantages of PTDE in such extreme conditions. Overall, the PTDE paradigm's ability to train a viable strategy with the assistance of agent-personalized global information and subsequently achieve decentralized execution through knowledge distillation is highlighted. The training curves (Figure \ref{sc2 curve} in Appendix C) and strategy visualizations (Appendix A) also illustrate the performance improvement of QMIX\_KD over QMIX.

\subsection{Google Research Football}

\begin{table*}[]\scriptsize 
\centering\caption{Winning rates on Google Research Football.}
\renewcommand{\arraystretch}{1.08} 
\begin{tabular}{lc@{\hspace{1mm}}|c@{\hspace{1mm}}c@{\hspace{1mm}}c@{\hspace{1mm}}c@{\hspace{1mm}}c}
\hline
\multicolumn{1}{c|}{}                                         & Algorithms          & \emph{3\_vs\_1\_w\_keeper}  & \emph{3\_vs\_2\_w\_keeper}  & \emph{counterattack\_easy} & \emph{counterattack\_hard}  & \emph{run\_pass\_and\_shoot\_w\_keeper}  \\ \hline
\multicolumn{1}{c|}{\multirow{2}{*}{\makecell[c]{Centralized \\ Execution}}} & QMIX\_GIU    & 0.662±0.256 & 0.415±0.201 & 0.839±0.075 & 0.462±0.240 & 0.687±0.072   \\ 
\multicolumn{1}{c|}{}                                         & \textbf{QMIX\_GIP (stage1)}    & \textbf{0.858±0.032} & \textbf{0.664±0.209} & \textbf{0.839±0.035} & \textbf{0.636±0.074} & \textbf{0.779±0.082} \\ \hline
\multicolumn{1}{c|}{\multirow{2}{*}{\makecell[c]{Decentralized \\ Execution}}}   & \textbf{QMIX\_KD (stage2)} & \textbf{0.818±0.056} & \textbf{0.732±0.138} & \textbf{0.734±0.176} & \textbf{0.517±0.053} & \textbf{0.775±0.055} \\ 
\multicolumn{1}{c|}{}                                         & QMIX         & 0.609±0.250 & 0.491±0.200 & 0.365±0.165 & 0.184±0.174 & 0.533±0.201   \\  \hline
\multicolumn{2}{c|}{Performance Retention Ratios (PRR)}   &  95.3\% &   110.2\%  &  87.5\%  &  81.3\%  &  99.5\%  \\ \hline
\end{tabular}
\label{grf winning rate}
\end{table*}

To further validate \textbf{RQ.1}, \textbf{RQ.2}, and \textbf{RQ.3}, we select five widely recognized academy scenarios: \emph{3\_vs\_1\_with\_keeper}, \emph{3\_vs\_2\_with\_keeper}, \emph{counterattack\_easy}, \emph{counterattack\_hard} and \emph{run\_pass\_and\_shoot\_with\_keeper}. The agents are trained for 10 million steps using 8 threads in all scenarios.

Table \ref{grf winning rate} displays winning rates on GRF. QMIX\_GIP achieves the best performance across all scenarios, reinforcing the notion that agent-personalized global information is more beneficial for multi-agent collaboration than unified global information (\textbf{RQ.2}). The high PRRs further demonstrate that the performance does not degrade significantly after knowledge distillation (\textbf{RQ.3}). Specifically, in \emph{3\_vs\_1\_with\_keeper} and \emph{run\_pass\_and\_shoot\_with\_keeper}, PRRs range from 95\% to 100\%, while in \emph{counterattack\_easy} and \emph{counterattack\_hard}, PRRs range from 80\% to 90\%. These conclusions align with the training curves available in Figure \ref{grf curve} in Appendix C. 


\subsection{Scenario Universality of PTDE Paradigm}

To further validate the effectiveness of the PTDE paradigm, we extend its application to the Learning to Rank (LTR) \cite{LTR} task. Ranking plays a crucial role in information retrieval, where the goal is to arrange a list of candidate documents in descending order of relevance to a given query. Achieving an optimal search ranking list enhances the effectiveness of information retrieval.

In the multi-agent cooperation setting for the Learning to Rank (LTR) task, each document is treated as an agent. The fundamental components of Multi-Agent Reinforcement Learning (MARL) are defined as follows:
\begin{itemize}
\setlength
\item Observation: Features of the query and document $i$.
\item State: Features of the query and all documents.
\item Reward: Given that NDCG@k \cite{jarvelin2002cumulated} is a standard evaluation metric for ranking, we define the reward as NDCG@k.
\item Action Space: Discrete scores, such as integers from 0 to 9, where the document is ultimately sorted based on the scores assigned to each document (agent).
\end{itemize}

\begin{figure}[t]
\centerline{\includegraphics[width=0.5\textwidth]{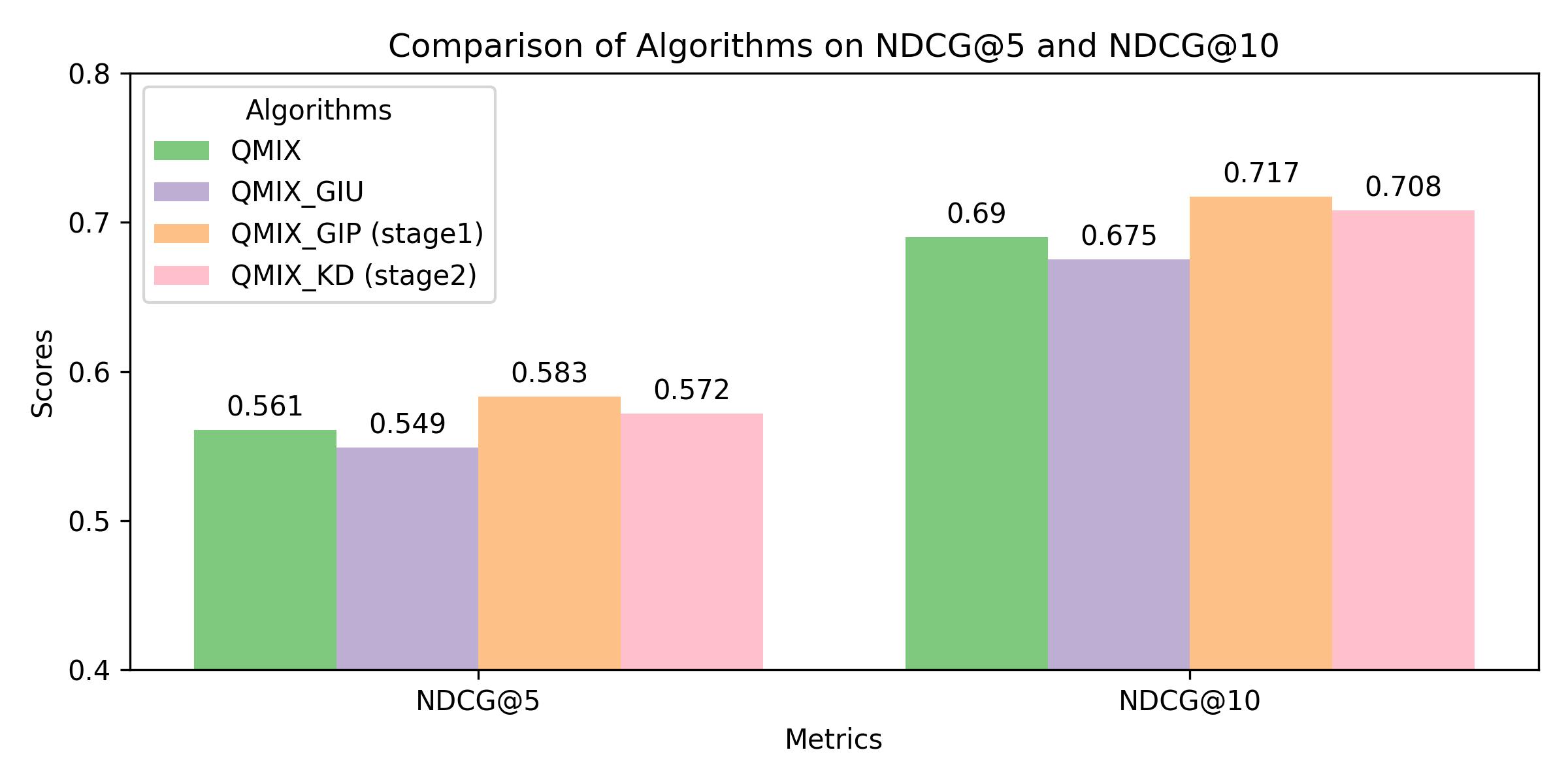}}
\caption{Experiments on Learning to Rank task.}
\label{bar}
\end{figure}

We conducted training and testing on 10,000 queries (7:3 partition) from the MSLR-WEB30K \cite{qin2013introducing} dataset, a large-scale dataset for Learning to Rank research. We modified this dataset to comprise 10 documents per query, resulting in a total of 10,000 queries and 100,000 documents. The experimental results are depicted in Figure \ref{bar}. Our approaches, QMIX\_GIP and QMIX\_KD, achieve higher NDCG scores compared to QMIX and QMIX\_GIU, addressing \textbf{RQ.2} and \textbf{RQ.3}. Interestingly, QMIX\_GIU performs worse than QMIX, providing additional insights into \textbf{RQ.1}.

Crucially, the experiments in LTR, along with those on the StarCraft II and GRF benchmarks, collectively demonstrate that the PTDE paradigm exhibits good universality across diverse scenarios, providing a comprehensive response to \textbf{RQ.4}.


\subsection{Algorithm Universality of PTDE Paradigm}

\definecolor{color1}{rgb}{0.95,0.95,0.95} 
\definecolor{color2}{rgb}{0.9,0.9,0.9}
\definecolor{color3}{rgb}{0.85,0.85,0.85}
\definecolor{color4}{rgb}{0.8,0.8,0.8} 

\begin{table}[] \scriptsize
\centering
\caption{Apply PTDE paradigm to VDN.}
\begin{tabular}{c|c|c}
\hline
Algorithm               & 3s\_vs\_5z (5M)  & 3s5z\_vs\_3s6z (10M) \\ \hline
VDN                     & \cellcolor{color1}0.653       & \cellcolor{color2}0.397          \\
VDN\_GIU                & \cellcolor{color3}0.971       & \cellcolor{color1}0.155          \\
VDN\_GIP (stage1)       & \cellcolor{color4}0.990       & \cellcolor{color4}0.704          \\
VDN\_KD (stage2)     & \cellcolor{color2}0.889       & \cellcolor{color3}0.609          \\ \hline
\end{tabular}
\label{PTDE_VDN}
\end{table}

\begin{table}[] \scriptsize
\centering
\caption{Apply PTDE paradigm to MAPPO.}
\begin{tabular}{c|c|c}
\hline
Algorithm               & 3s\_vs\_5z (5M)  & 3s5z\_vs\_3s6z (10M) \\ \hline
MAPPO                   & \cellcolor{color2}0.767       & \cellcolor{color1}0.000          \\
MAPPO\_GIU              & \cellcolor{color1}0.602       & \cellcolor{color2}0.486          \\
MAPPO\_GIP (stage1)     & \cellcolor{color4}0.965       & \cellcolor{color4}0.694          \\
MAPPO\_KD (stage2)   & \cellcolor{color3}0.891       & \cellcolor{color3}0.589          \\ \hline
\end{tabular}
\label{PTDE_MAPPO}
\end{table}

In this section, we integrate the PTDE paradigm with VDN and MAPPO, and test them on the \emph{3s\_vs\_5z} and \emph{3s5z\_vs\_3s6z} scenarios. We adopt hyperparameters as specified in \cite{PYMARL2} and \cite{MAPPO}, respectively. As shown in Table \ref{PTDE_VDN} and \ref{PTDE_MAPPO}, VDN\_GIU's performance on \emph{3s5z\_vs\_3s6z} is inferior to that of VDN, and MAPPO\_GIU's performance on \emph{3s\_vs\_5z} is worse than MAPPO. This highlights that unified global information does not always contribute to the efficacy of multi-agent collaboration (\textbf{RQ.1}). Furthermore, VDN\_GIP outperforms VDN\_GIU, and MAPPO\_GIP outperforms MAPPO\_GIU in both scenarios, demonstrating the effectiveness of the PTDE paradigm for both value-decomposition-based algorithm VDN and actor-critic-based algorithm MAPPO. In essence, the PTDE paradigm exhibits good universality across different algorithm types (\textbf{RQ.5}). The training curves for Table \ref{PTDE_VDN} and Table \ref{PTDE_MAPPO} can be found in Figure \ref{Universality} in Appendix C.



\subsection{Empirical Analysis of the Two-Stage Training}

Why is it necessary to divide the training process into two stages? In Table \ref{CTDS_experiment}, we compare the PTDE and CTDS paradigms based on two metrics: winning rate and PRR. CTDS synchronizes the distillation of global policies with centralized training, while our PTDE approach first conducts centralized training and then performs agent-personalized global knowledge distillation. Across the \emph{5m\_vs\_6m}, \emph{6h\_vs\_8z}, and \emph{3s5z\_vs\_3s7z} maps, the PRR metric consistently favors the PTDE paradigm over the CTDS paradigm. This underscores that the two-stage training approach of PTDE maintains superior performance during decentralized execution (addressing \textbf{RQ.3} and \textbf{RQ.6}).



\begin{table}[t] \scriptsize
\centering
\caption{Comparisons between PTDE and CTDS.}
\begin{tabular}{c|ccc}
\hline
Algorithm                           & 5m\_vs\_6m  & 6h\_vs\_8z & 3s5z\_vs\_3s7z \\ \hline
QMIX\_GIP (stage1)                 & 0.806       & 0.712      & 0.710          \\ 
QMIX\_KD (stage2)               & 0.690       & 0.524      & 0.631          \\
PRR   & 85.6\%      & 73.6\%     & 88.9\%         \\ \hline
CTDS (QMIX\_Teacher)                & 0.698       & 0.367      & 0.000          \\ 
CTDS (QMIX\_Student)                & 0.490       & 0.204      & 0.000          \\
PRR   & 70.2\%      & 55.6\%     & -           \\ \hline
\end{tabular}
\label{CTDS_experiment}
\end{table}

\begin{figure}[t]
    \centerline{\includegraphics[width=0.5\textwidth]{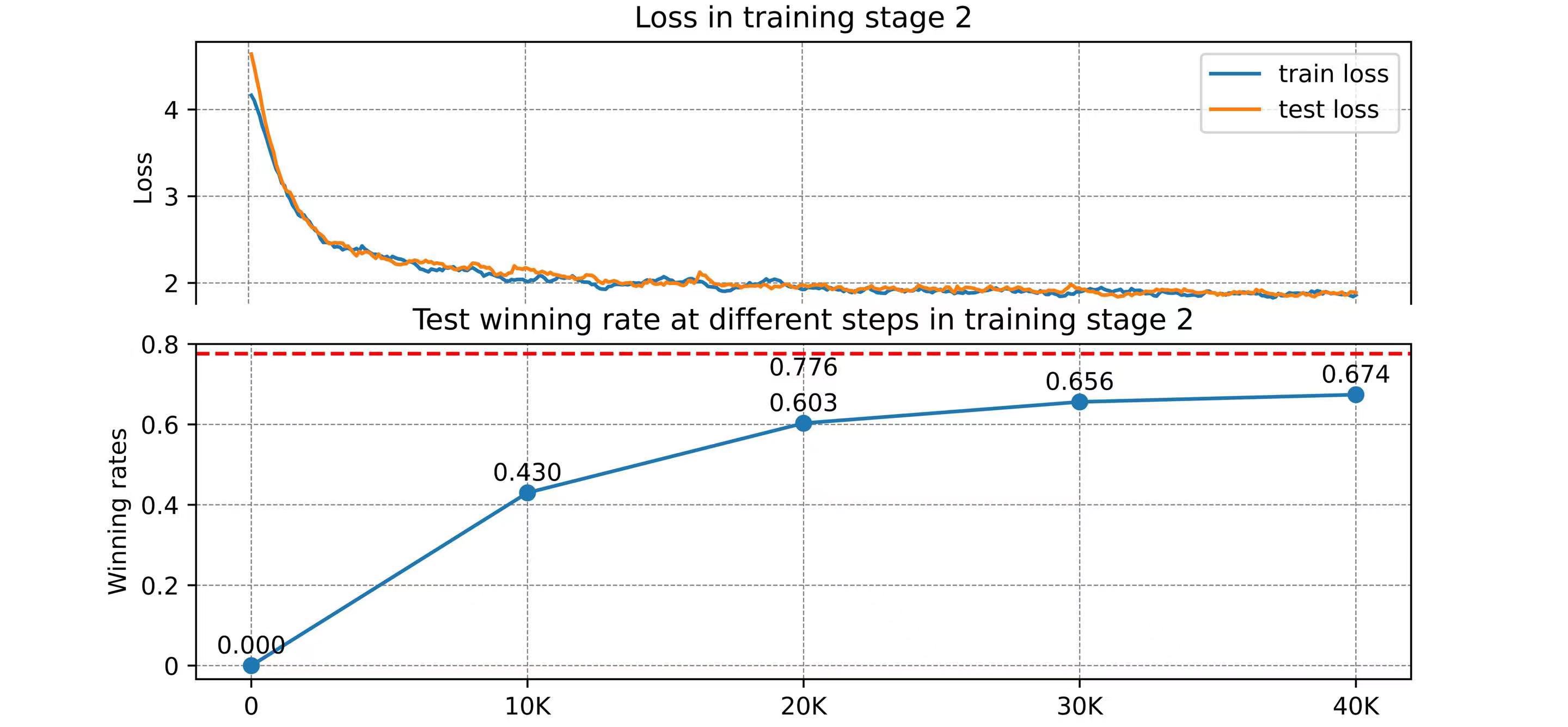}}
    \caption{The loss and winning rates in training stage 2.}
    \label{stage2_curve}
\end{figure}

Figure \ref{stage2_curve} displays the loss function (\ref{MSE}) and test winning rates during the knowledge distillation process on the \emph{3s\_vs\_5z} scenario. The loss plot displays training and testing loss curves spanning 0 to 40k steps. The winning rate plot showcases a blue line representing test winning rates for decentralized execution after knowledge distillation, and a red dashed line signifying the test winning rate for centralized execution before knowledge distillation. In this experiment, the batch size is set to 1000, and each step corresponds to the training of one batch. As can be observed, a notable reduction in loss occurs before 30k steps, accompanied by a swift increase in the test winning rate. This suggests that distilling global policies using local information requires a specific training duration. Conversely, simultaneous knowledge distillation during centralized training might fail to preserve optimal performance in decentralized execution due to inadequate training and uncertainties in the distribution of training samples. This analysis addresses and responds to \textbf{RQ.6}.

\section{Conclusions}

We introduced a two-stage training paradigm, named Personalized Training with Distilled Execution (PTDE), designed for multi-agent reinforcement learning. In the first training stage, the Global Information Personalization (GIP) module tailors global information for each agent. Subsequently, during the second training stage, the student network distills agent-personalized global information using solely the local information of each agent. In the execution stage, the student network takes over from the teacher network, enabling decentralized execution.

Our empirical evaluations on the StarCraft II benchmark, Google Research Football benchmark, and Learning to Rank task collectively offer conclusive answers to the posed research questions (RQ.1 to RQ.6). These results provide robust evidence supporting the efficacy and broad applicability of the PTDE paradigm, indicating its promise within the multi-agent reinforcement learning domain.


\clearpage

\section*{Acknowledgments}

This research was supported by the Natural Science Foundation of China (61902209, 62377044, U2001212), and Beijing Outstanding Young Scientist Program (NO.BJJWZYJH012019100020098) and Intelligent Social Governance Platform, Major Innovation \& Planning Interdisciplinary Platform for the "Double-First Class” Initiative, Renmin University of China.

\bibliographystyle{named}
\bibliography{ijcai24}

\clearpage

\appendix
\begin{center}
\textbf{\LARGE Appendix}
\end{center}


\section{Strategy Visualization}
\begin{figure}[htbp]
\centering
    \subfigure[{\scriptsize QMIX (early stage)}]{
    \begin{minipage}[t]{0.22\textwidth}
    \includegraphics[width=1\textwidth]{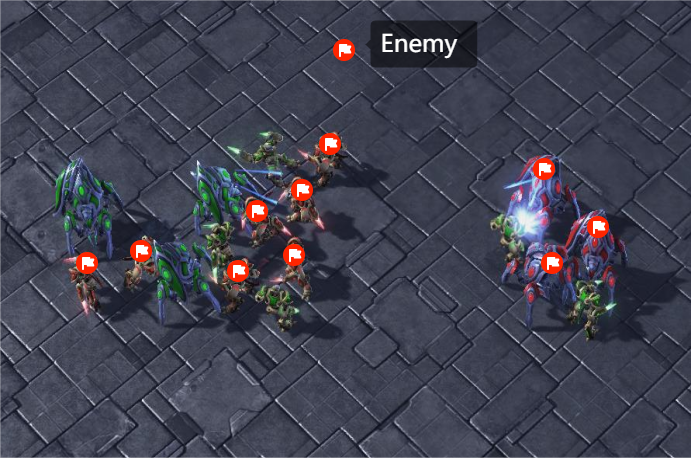}
    \end{minipage}
    }
    \subfigure[{\scriptsize QMIX (late stage)}]{
    \begin{minipage}[t]{0.22\textwidth}
    \includegraphics[width=1\textwidth]{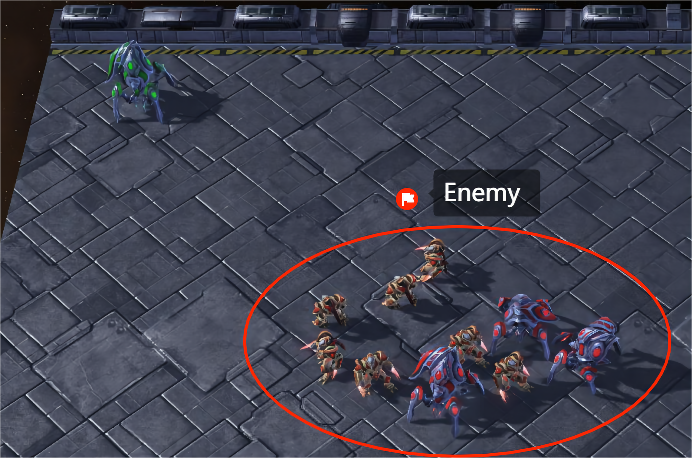}
    \end{minipage}
    }
    \subfigure[{\scriptsize QMIX\_KD (early stage)}]{
    \begin{minipage}[t]{0.22\textwidth}
    \includegraphics[width=1\textwidth]{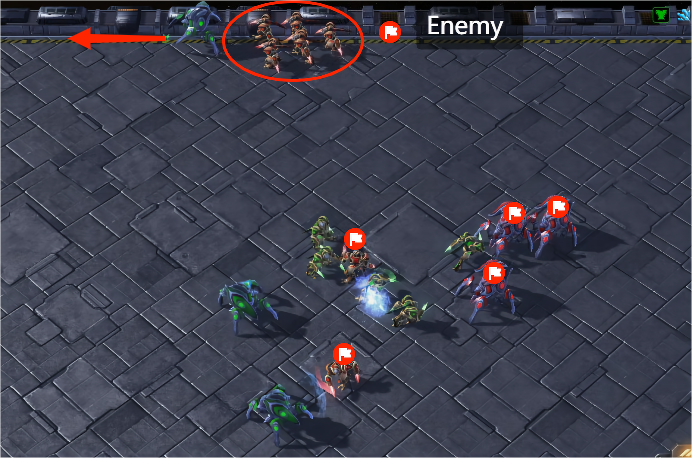}
    \end{minipage}
    }
    \subfigure[{\scriptsize QMIX\_KD (late stage)}]{
    \begin{minipage}[t]{0.22\textwidth}
    \includegraphics[width=1\textwidth]{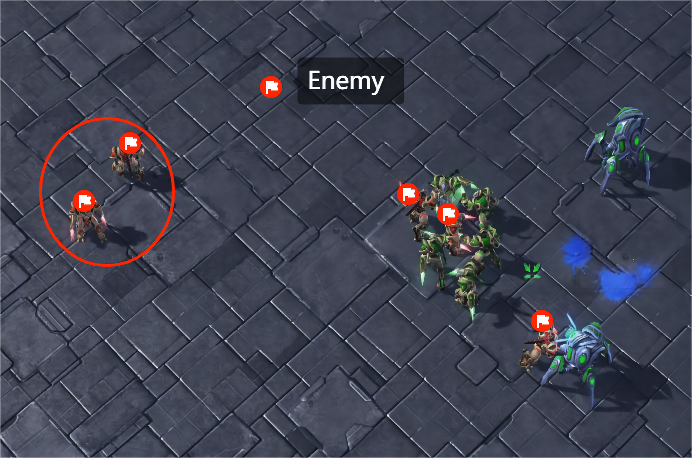}
    \end{minipage}
    }
\caption{The Well-Trained Strategy Visualization: QMIX and QMIX\_KD on \emph{3s5z\_vs\_3s7z} scenario.}
\label{video}
\end{figure}

To further figure out the performance improvement brought by PTDE paradigm, \textbf{Strategy Visualization} is shown in Figure \ref{video}, which is strategy comparison of well trained QMIX\_KD and QMIX on \emph{3s5z\_vs\_3s7z} scenario. Subfigures (a) and (b) show QMIX's strategies in the early and late stage of an episode, respectively. As shown in subfigure (a), in the early stage of the episode, QMIX's strategy is to play head-on against the enemies. However, due to the larger number of enemies, only one agent unit left in the late stage of the episode as shown in subfigure (b). Finally, the agents controlled by QMIX are defeated by the enemy. Subfigures (c) and (d) show QMIX\_KD's strategies. From subfigure (c), we can see that there is a Stalker which takes advantage of its fast moving speed to attract five enemy's Zealots by itself in the early stage. This allows the other agents to form a local advantage in unit number, killing the enemy's three Stalkers and two Zealots first. When the episode reaches the later stage, our agents use a new strategy to defeat the five enemy's Zealots which are attracted by the Stalker earlier. As shown in subfigure (d), agents attract three enemy's Zealots for quick killing. Then, the last two enemy's Zealots will also be killed by agents due to the disadvantage of smaller unit number. 

\section{Further Explanation of Two-Stage Training}

Why we use two-stage training rather than simultaneous training of knowledge distillation while interacting with the environment? There are two main reasons: 

\begin{itemize}
\item First, during the first training stage, the parameters of teacher network are dynamically changing. So the distribution of $z_i^t$ is unstable, which does not meet the identical distribution requirement of supervised learning.
\item Second, for some on-policy MARL algorithms based on actor-critic, such as MAPPO, the data used to update network parameters represents the current moment data. However, for some off-policy algorithms based on value-decomposition, such as QMIX, the data used for network updating is the past data sampled from buffer. The distribution of these data does not represent the parameters of teacher network at the current moment. This exacerbates the heterogeneous distribution of data.
\end{itemize}

In contrast, after the first stage of training, the distribution of target values $z_i^t$ in the second training stage is fixed, which is more friendly for knowledge distillation. So we choose a two-stage training strategy. In experiments (Section 4.5), We also prove that it's necessary to follow two-stage training paradigm.

\section{Curves and Tables}

\begin{figure*}[h]
\centerline{\includegraphics[width=1\textwidth]{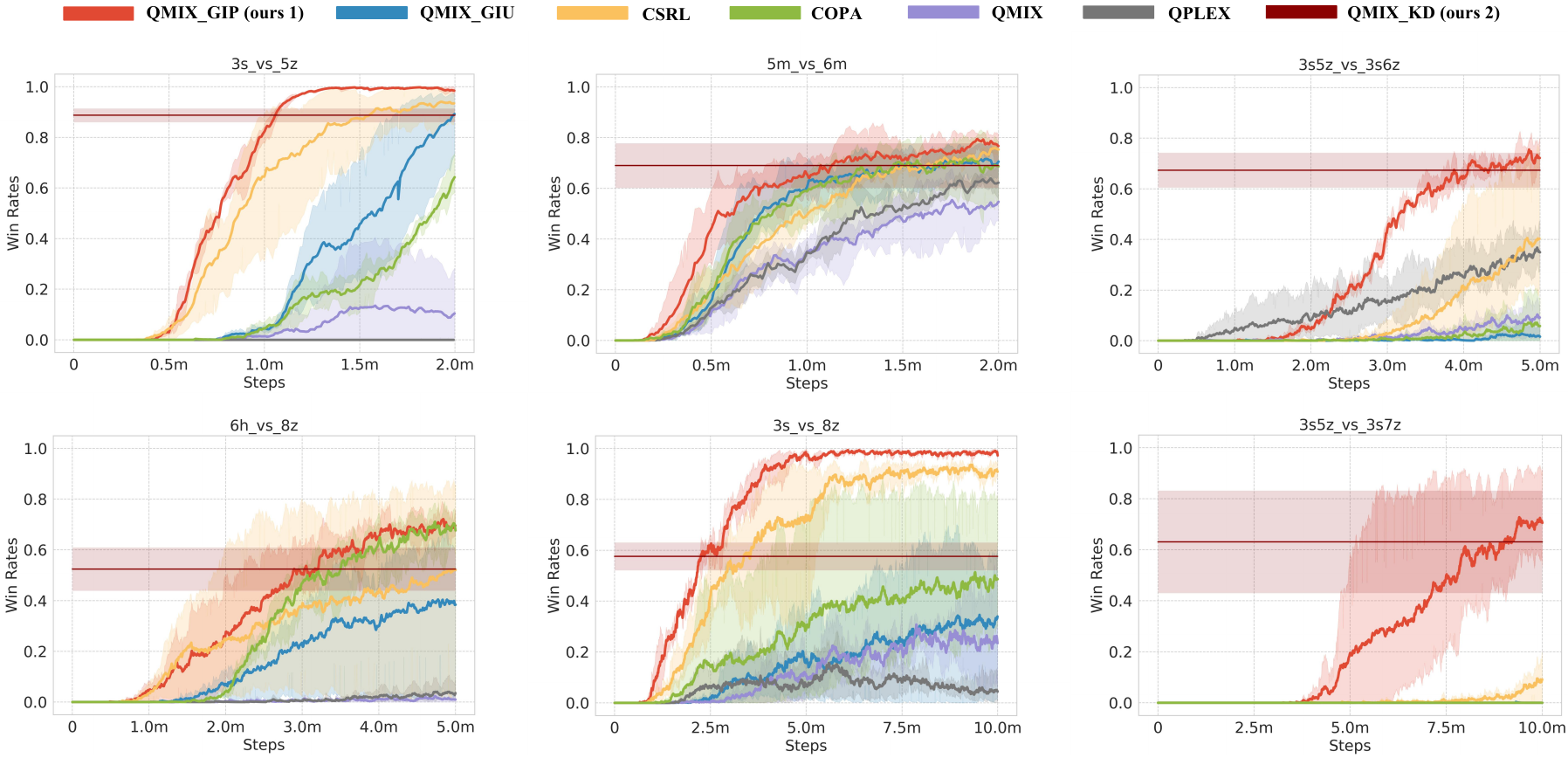}}
\caption{Experiments on StarCraft II benchmark. QMIX\_GIP is the method after the first training stage, QMIX\_KD is the method after the second training stage. QMIX\_GIU is the method which uses unified global information during execution. CSRL and COPA are two centralized execution algorithms. QMIX and QPLEX are two decentralized algorithms.}
\label{sc2 curve}
\end{figure*}

\begin{figure*}[h]
\centerline{\includegraphics[width=1\textwidth]{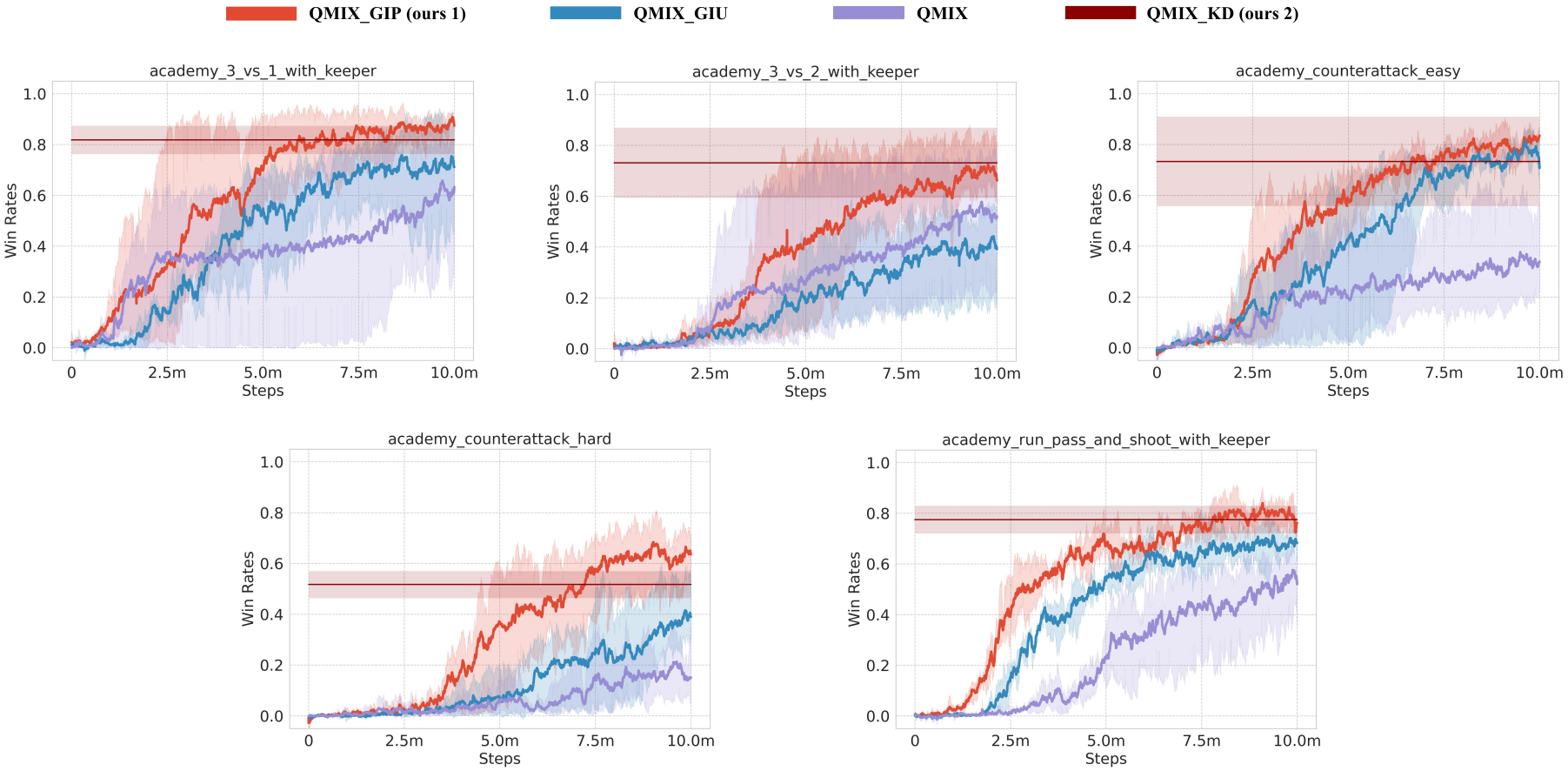}}
\caption{Experiments on Google Research Football benchmark}
\label{grf curve}
\end{figure*}

\begin{figure*}[thbp!]
\centering
    \subfigure[\emph{3s\_vs\_5z}\_VDN]{
    \begin{minipage}[t]{0.45\textwidth}
    \includegraphics[width=1\textwidth]{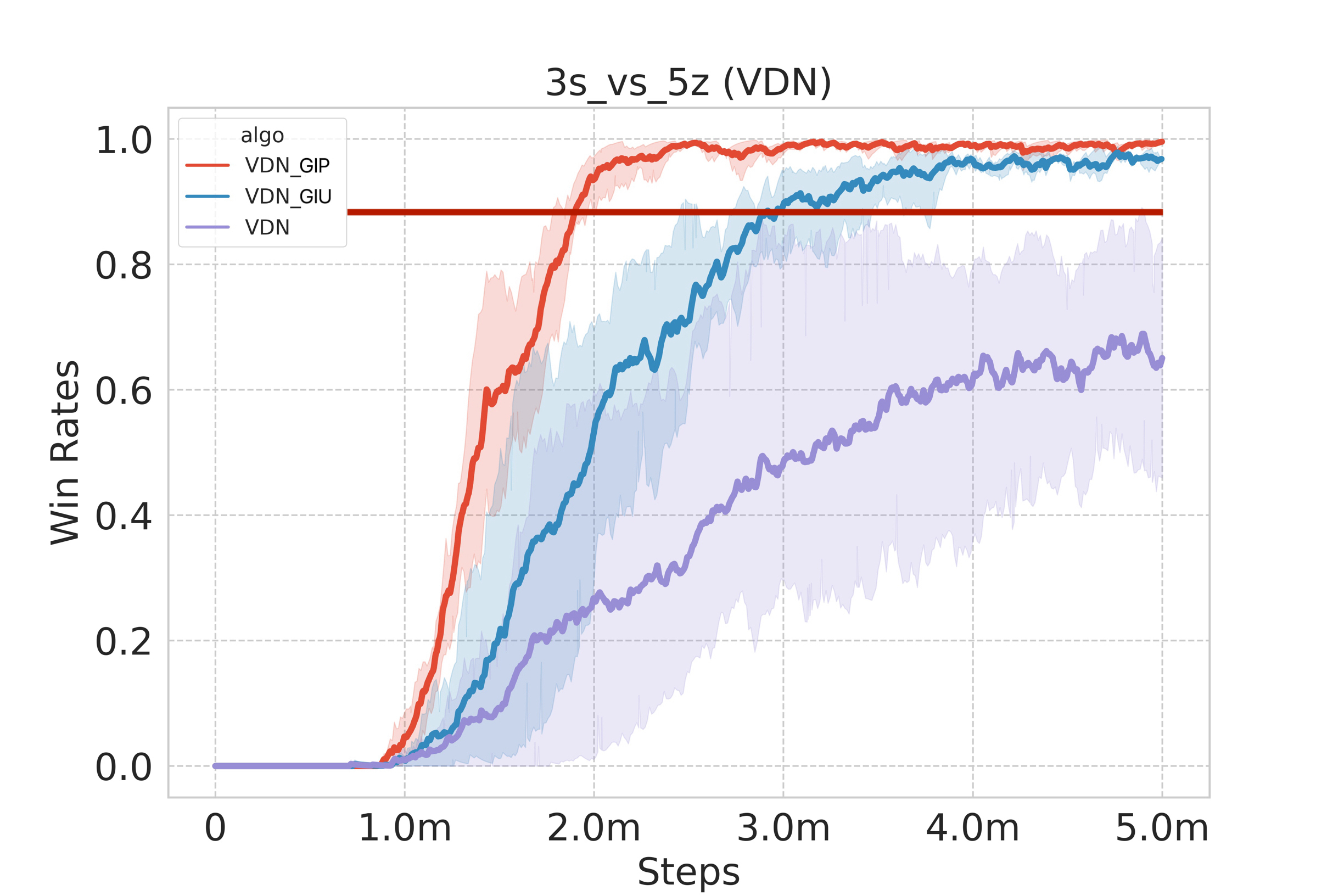}
    \end{minipage}
    }
    \subfigure[\emph{3s5z\_vs\_3s6z}\_VDN]{
    \begin{minipage}[t]{0.45\textwidth}
    \includegraphics[width=1\textwidth]{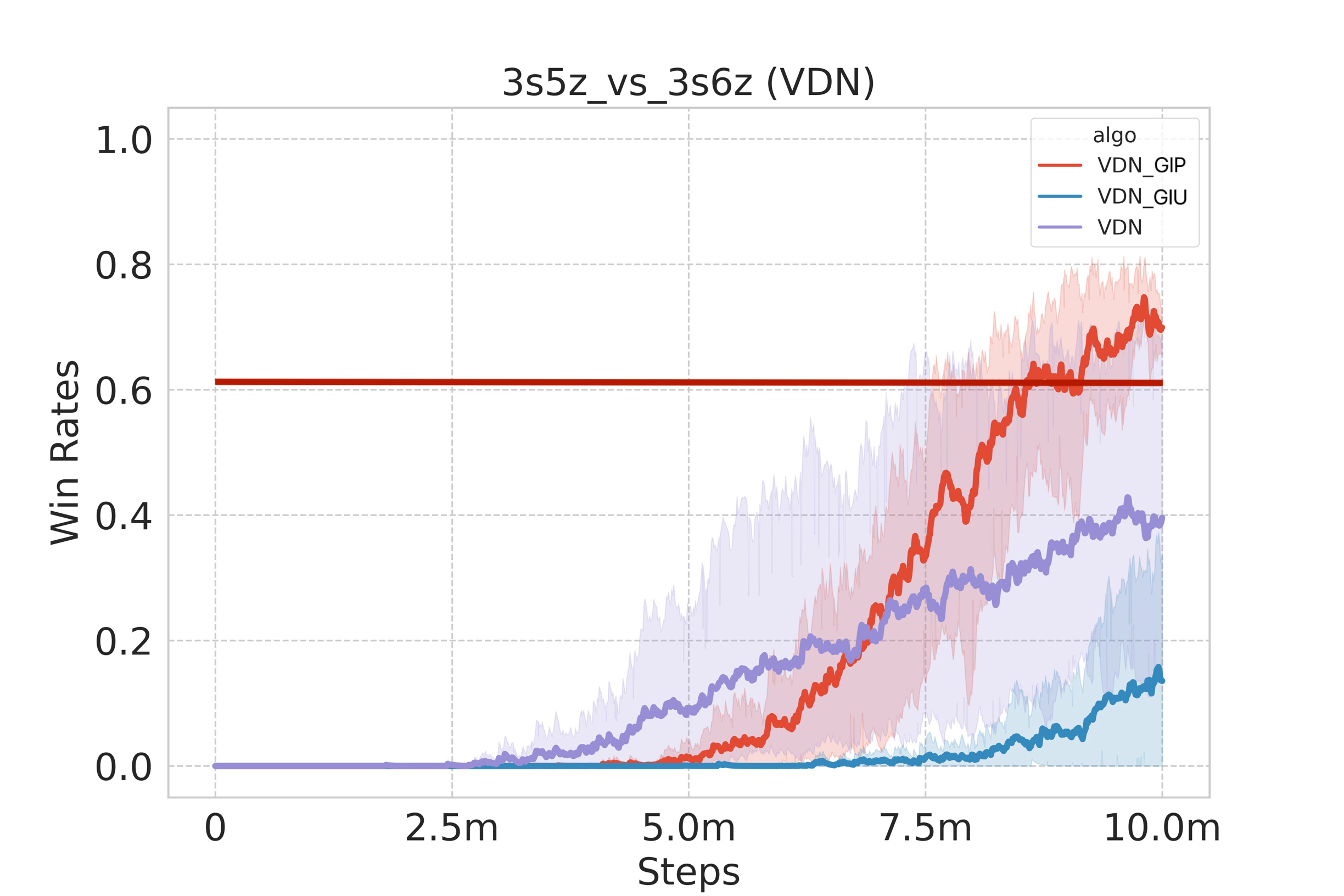}
    \end{minipage}
    }
    \subfigure[\emph{3s\_vs\_5z}\_MAPPO]{
    \begin{minipage}[t]{0.45\textwidth}
    \includegraphics[width=1\textwidth]{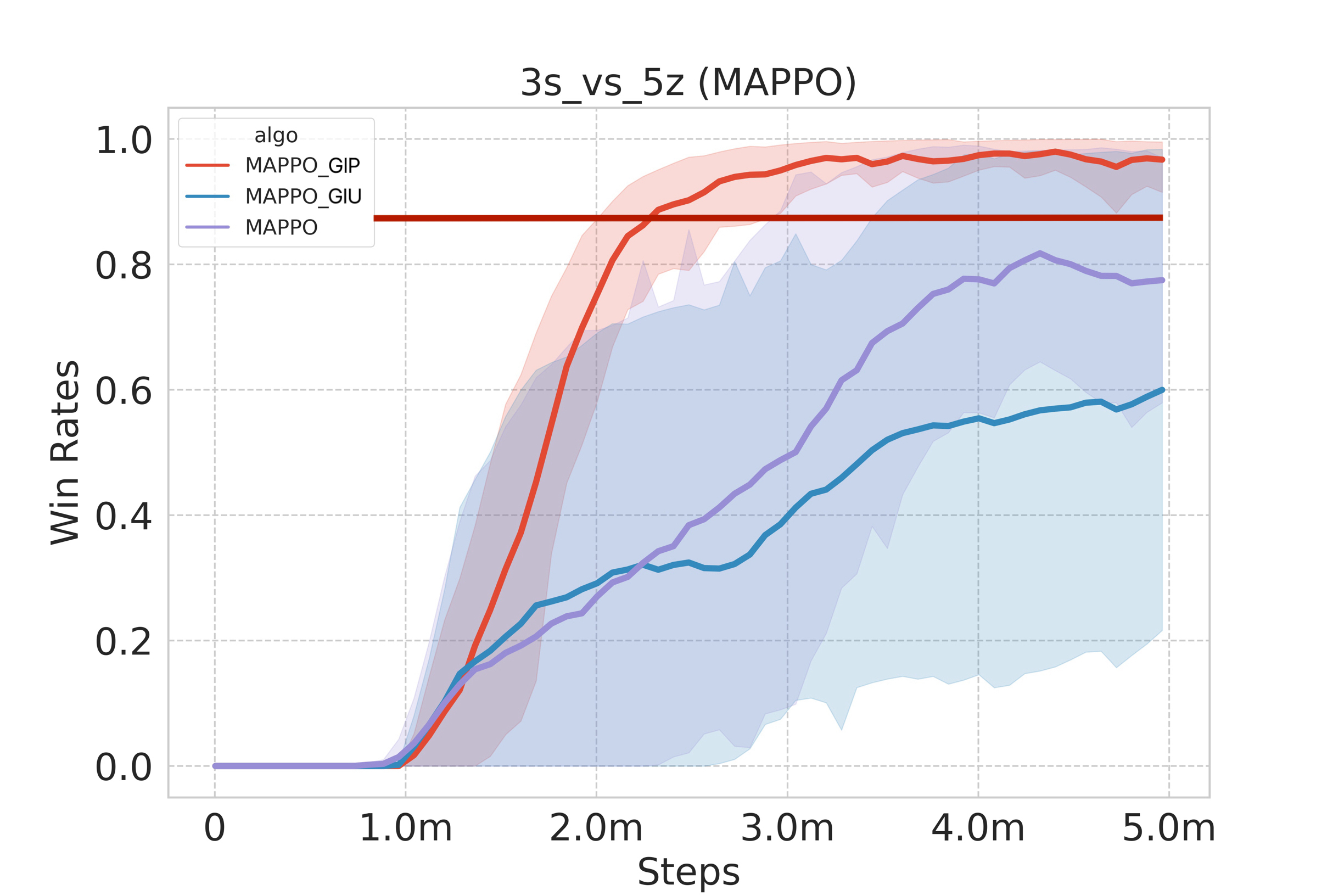}
    \end{minipage}
    }
    \subfigure[\emph{3s5z\_vs\_3s6z}\_MAPPO]{
    \begin{minipage}[t]{0.45\textwidth}
    \includegraphics[width=1\textwidth]{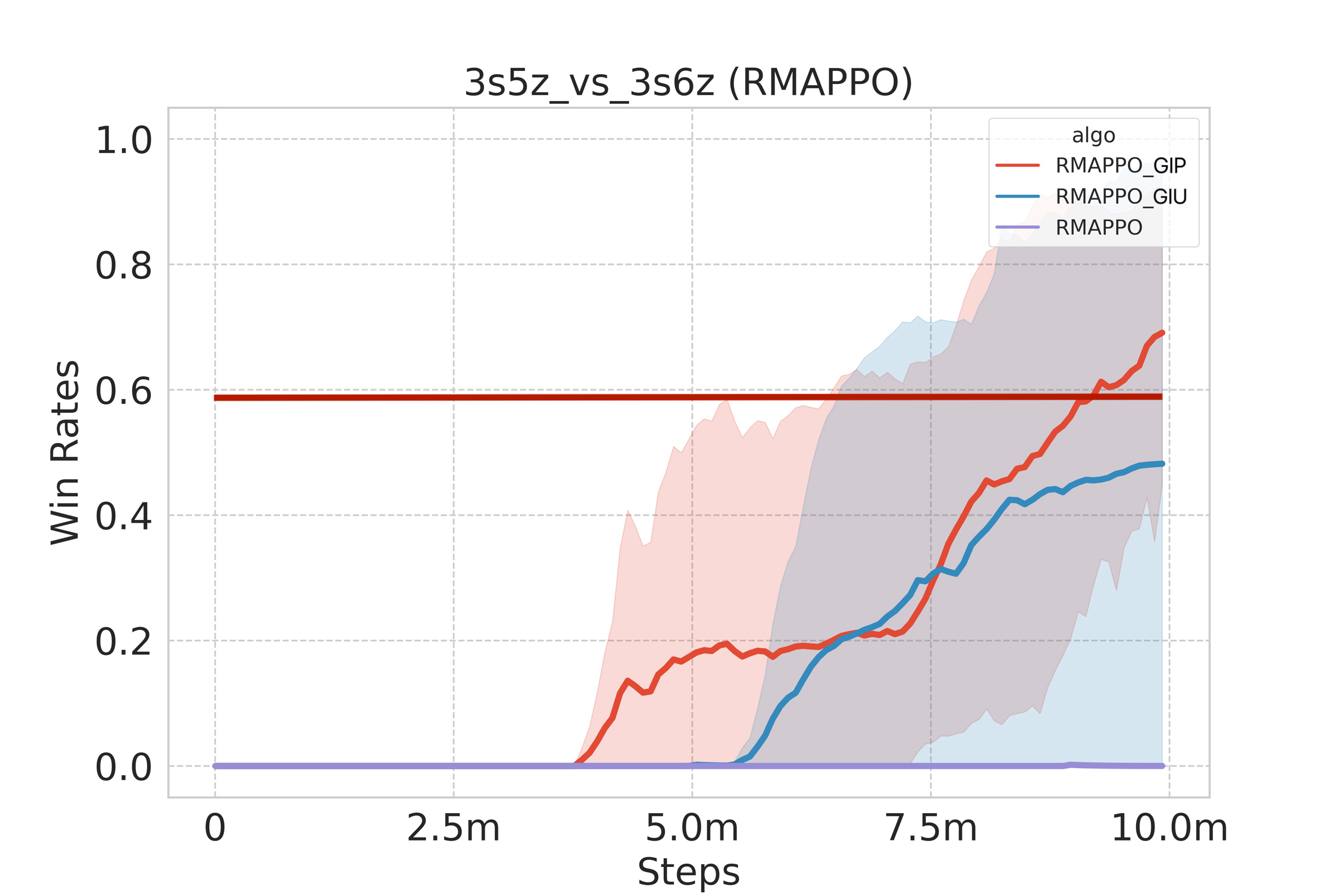}
    \end{minipage}
    }
\caption{Universality Experiments on SMAC. The dark red line in the graph represents the average win rate after KD.}
\label{Universality}
\end{figure*}

The next content of Appendix primarily includes experimental curves for StarCraft II (Figure \ref{sc2 curve}) and Google Research Football (Figure \ref{grf curve}), the universality experiments on SMAC (Figure \ref{Universality}), as well as the hyperparameters settings for the algorithms (Table \ref{hyper parameters}) and the feature engineering for Google Research Football (Table \ref{grf feature}).


\clearpage

\clearpage

\begin{table*}[] 
\centering\caption{Hyperparemeters Settings of value-decomposition based algorithms. For fairness of experiments, the hyperparameters of all algorithms on all maps are same, including epsilon\_anneal\_time (500000 is best for \emph{6h\_vs\_8z} in PyMARL2), batch\_size\_run (4 is best for \emph{3s5z\_vs\_3s6z} in PyMARL2), batch\_size (64 is best for \emph{3s5z\_vs\_3s6z} in PyMARL2), buffer\_size (2500 is best for \emph{3s5z\_vs\_3s6z} in PyMARL2) and td\_lambda (0.3 is best for \emph{6h\_vs\_8z} in PyMARL2).}
\begin{tabular}{|ll|}
\hline
\multicolumn{1}{|c|}{Parameter Name}           & Value                \\ \hline
\multicolumn{2}{|c|}{Exploration-related}                             \\ \hline
\multicolumn{1}{|l|}{action-selector}          & epsilon\_greedy      \\ \hline
\multicolumn{1}{|l|}{epsilon\_start}           & 1.0                  \\ \hline
\multicolumn{1}{|l|}{epsilon\_finish}          & 0.05                 \\ \hline
\multicolumn{1}{|l|}{epsilon\_anneal\_time}    & 100000 (for all maps) \\ \hline
\multicolumn{2}{|c|}{Sampler-related}                                 \\ \hline
\multicolumn{1}{|l|}{runner}                   & parallel             \\ \hline
\multicolumn{1}{|l|}{batch\_size\_run}         & 8 (for all maps)     \\ \hline
\multicolumn{1}{|l|}{batch\_size}              & 128 (for all maps)   \\ \hline
\multicolumn{1}{|l|}{buffer\_size}             & 5000 (for all maps)  \\ \hline
\multicolumn{1}{|l|}{t\_max}                   & 10000000             \\ \hline
\multicolumn{2}{|c|}{Agent-related}                                   \\ \hline
\multicolumn{1}{|l|}{rnn\_hidden\_dim}         & 64                   \\ \hline
\multicolumn{1}{|l|}{hyper\_hidden\_dim}       & 64                   \\ \hline
\multicolumn{1}{|l|}{z\_dim}                   & 64                   \\ \hline
\multicolumn{2}{|c|}{Training-related}                                \\ \hline
\multicolumn{1}{|l|}{mixer}                    & qmix or dqn          \\ \hline
\multicolumn{1}{|l|}{lr}                       & 0.001                \\ \hline
\multicolumn{1}{|l|}{td\_lambda}               & 0.6 (for all maps)    \\ \hline
\multicolumn{1}{|l|}{optimizer}                & adam                 \\ \hline
\multicolumn{1}{|l|}{target\_update\_interval} & 200                  \\ \hline
\end{tabular}
\label{hyper parameters}
\end{table*}


\begin{table*}[] 
\centering\caption{The feature composition of observation and state in Google Research Football}
\begin{tabular}{|c|c|c|}
\hline
\multirow{8}{*}{Observation} & \multirow{2}{*}{Player}     & Absolute position     \\ \cline{3-3} 
                             &                             & Absolute speed        \\ \cline{2-3} 
                             & \multirow{2}{*}{Left team}  & Relative position     \\ \cline{3-3} 
                             &                             & Relative speed        \\ \cline{2-3} 
                             & \multirow{2}{*}{Right team} & Relative position     \\ \cline{3-3} 
                             &                             & Relative speed        \\ \cline{2-3} 
                             & \multirow{2}{*}{Ball}       & Absolute position     \\ \cline{3-3} 
                             &                             & Belong to (team ID)   \\ \hline
\multirow{13}{*}{State}      & \multirow{4}{*}{Left team}  & Absolute position     \\ \cline{3-3} 
                             &                             & Absolute speed        \\ \cline{3-3} 
                             &                             & Tired factor          \\ \cline{3-3} 
                             &                             & Player type           \\ \cline{2-3} 
                             & \multirow{4}{*}{Right team} & Absolute position     \\ \cline{3-3} 
                             &                             & Absolute speed        \\ \cline{3-3} 
                             &                             & Tired factor          \\ \cline{3-3} 
                             &                             & Player type           \\ \cline{2-3} 
                             & \multirow{5}{*}{Ball}       & Absolute position     \\ \cline{3-3} 
                             &                             & Absolute speed        \\ \cline{3-3} 
                             &                             & Absolute rotate speed \\ \cline{3-3} 
                             &                             & Belong to (team ID)   \\ \cline{3-3} 
                             &                             & Belong to (player ID) \\ \hline
\end{tabular}
\label{grf feature}
\end{table*}

\end{document}